%% file: main.tex
\documentclass[10pt,twocolumn,letterpaper]{article}

\usepackage{cvpr}              
\usepackage[accsupp]{axessibility} 

\input{preamble}
\definecolor{cvprblue}{rgb}{0.21,0.49,0.74}
\usepackage[pagebackref,breaklinks,colorlinks,allcolors=cvprblue]{hyperref}

\def\paperID{} 
\def\confName{CVPR}
\def\confYear{2026}

\title{MatPedia: A Universal Generative Foundation for High-Fidelity Material Synthesis}

\author{
Di Luo\textsuperscript{1,2*} \quad
Shuhui Yang\textsuperscript{2*} \quad
Mingxin Yang\textsuperscript{2*} \quad
Jiawei Lu\textsuperscript{1} \quad
Yixuan Tang\textsuperscript{2,3}  \\
Xintong Han\textsuperscript{2} \quad
Zhuo Chen\textsuperscript{2} \quad
Beibei Wang\textsuperscript{4$\dagger$} \quad
Chunchao Guo\textsuperscript{2$\dagger$}
\\[0.5em]
\textsuperscript{1}Nankai University \quad
\textsuperscript{2}Tencent Hunyuan \quad
\textsuperscript{3}Xi'an Jiaotong University \quad
\textsuperscript{4}Nanjing University
\\[0.5em]
\texttt{\small \{diluo, jsljiawei\}@mail.nankai.edu.cn}, 
\texttt{\small beibei.wang@nju.edu.cn} \\
\texttt{\small \{oakyang,maddoxyang,setskytang,pathan,zhuooochen,chunchaoguo\}@tencent.com}\\
}

\begin{document}



\twocolumn[{
\renewcommand\twocolumn[1][]{#1}
\maketitle
\vspace{-9mm}

\begin{center}
\vspace{-10pt}
    \captionsetup{type=figure}
    \includegraphics[width=1.0\textwidth]{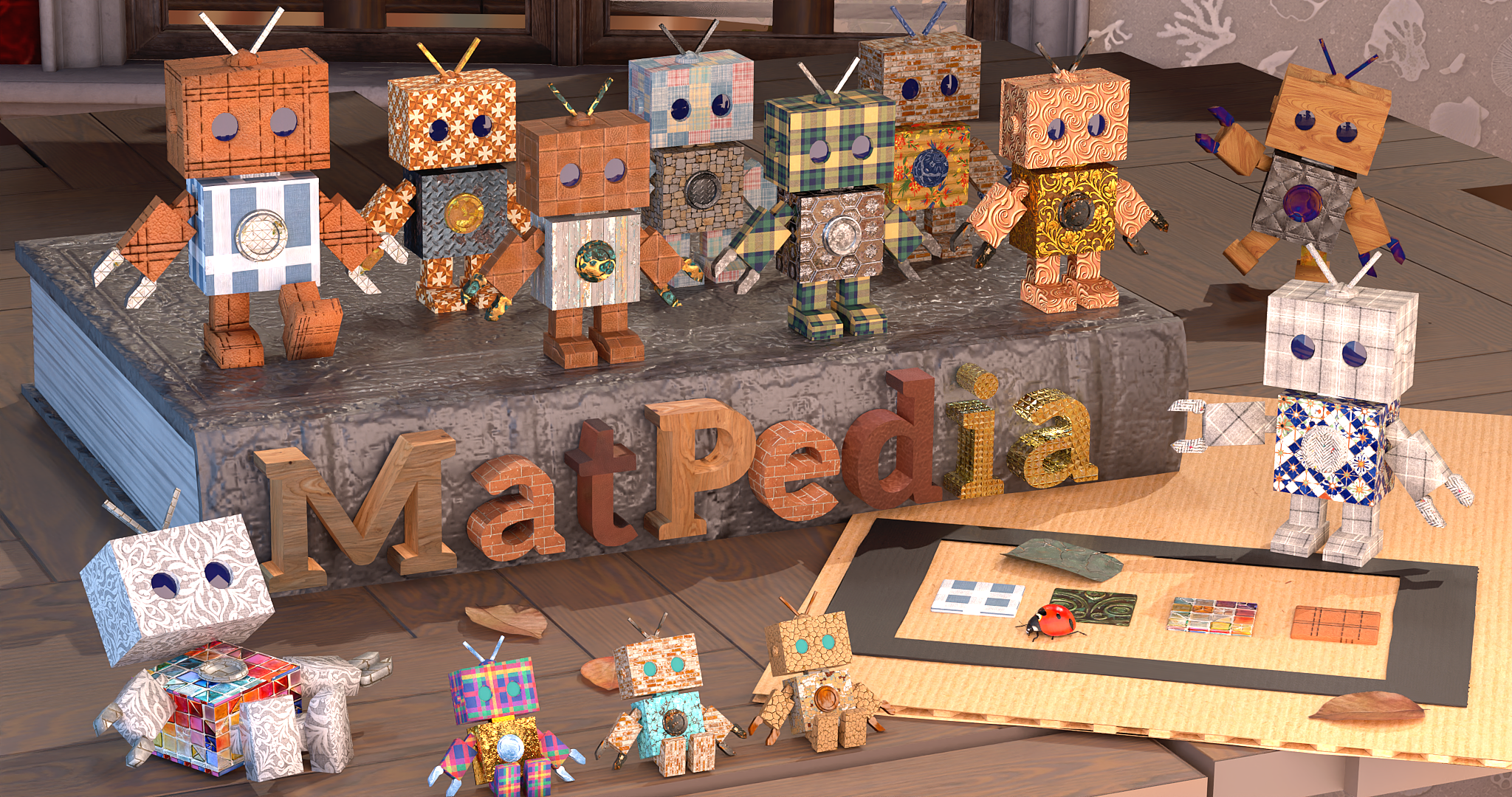}
    \vspace{-18pt}
    \caption{We introduce \textbf{MatPedia}, a foundation model with a joint RGB-PBR representation that supports diverse material-related tasks. A vibrant ensemble of robots and the ``MatPedia" typography, each adorned with distinct, photorealistic materials entirely synthesized by our method, showcasing our work’s capability in generating diverse, high-fidelity materials for 3D assets.}
    \label{fig:teaser}
    \vspace{-5pt}
\end{center}
}]

\def\thefootnote{}\footnotetext{* Equal Contribution.} \def\thefootnote{}\footnotetext{$\dagger$ Corresponding author.}

\input{sec/0_abstract}

\input{sec/1_intro}
\input{sec/2_related_work}

\input{sec/3_method}

\input{sec/4_experiment}
{
    \small
    \bibliographystyle{ieeenat_fullname}
    \bibliography{main}
}


\end{document}

%% file: sec/0_abstract.tex
\begin{abstract}

Physically-based rendering (PBR) materials are fundamental to photorealistic 
graphics, yet their creation remains labor-intensive and requires specialized 
expertise. While generative models have advanced material synthesis, existing 
methods lack a unified representation bridging natural image appearance and 
PBR properties, leading to fragmented task-specific pipelines and inability 
to leverage large-scale RGB image data. We present \textbf{MatPedia}, a 
foundation model built upon a novel joint RGB-PBR representation that compactly 
encodes materials into two interdependent latents: one for RGB appearance and 
one for the four PBR maps encoding complementary physical properties. By formulating 
them as a 5-frame sequence and employing video diffusion architectures, 
MatPedia naturally captures their correlations while transferring visual priors 
from RGB generation models. This joint representation enables a unified framework 
handling multiple material tasks—text-to-material generation, image-to-material 
generation, and intrinsic decomposition—within a single architecture. Trained 
on \textbf{MatHybrid-410K}, a mixed corpus combining PBR datasets with large-scale 
RGB images, MatPedia achieves native 1024×1024 synthesis that substantially 
surpasses existing approaches in both quality and diversity.
\end{abstract}

%% file: sec/1_intro.tex
\section{Introduction}
\label{sec:intro}

Physically-based rendering materials are fundamental to modern computer graphics, enabling the creation of photorealistic surfaces that respond accurately to diverse lighting and viewing conditions. They are essential in applications such as visual effects, video games, virtual reality, architectural visualization, and industrial design. As rendering technologies advance—supporting higher resolutions and more complex scenes—the demand for diverse, high-quality material libraries grows substantially. However, creating these materials remains a meticulous, labor-intensive process that requires specialized expertise and tools, thereby limiting the scalability of asset production.

The extension of generative models, notably GANs~\cite{goodfellow2020generative} 
and diffusion models~\cite{ho2020denoising,rombach2022high}, to PBR material 
synthesis has enabled the generation of complex, interdependent texture maps 
(such as basecolor, metallic, roughness, and normal) from various inputs~\cite{vecchio2024controlmat,zhou2022tilegen,vecchio2024matfuse,ma2024materialpicker,lopes2024material,zeng2024rgb}. 
However, current approaches lack a unified latent representation bridging 
natural image appearance (RGB) and PBR material properties. This fundamental gap 
leads to two critical limitations. First, existing methods remain fragmented 
into task-specific pipelines—such as intrinsic decomposition~\cite{he2025materialmvp,liang2025diffusion,zeng2024rgb} 
versus direct material generation~\cite{vecchio2024controlmat,vecchio2024matfuse,ma2024materialpicker}—preventing 
universal architectures capable of handling diverse material-related tasks. 
Second, they are constrained to training on limited PBR datasets~\cite{vecchio2024matsynth,ma2023opensvbrdf,deschaintre2018single}, 
unable to leverage large-scale, high-quality RGB data. Consequently, the 
quality and diversity of synthesized materials remain far below the potential 
demonstrated by modern RGB image generators.

To address these challenges, we introduce \textbf{MatPedia}, a foundation model based on a \textit{joint RGB-PBR representation} that unifies visual appearance and physical material properties within a single encoding framework. This representation supports universal architectures for diverse material tasks and enables the use of large-scale RGB image data for material synthesis. Our key observation is that RGB images already contain rich cues about material appearance, while PBR maps provide the underlying physical explanations of that appearance. Leveraging this asymmetry, we encode PBR maps conditioned on the RGB image rather than treating them as independent modalities. The four PBR maps (basecolor, metallic, roughness, normal) collectively describe the physical properties manifested in the RGB image; thus, they can be compressed into a single latent vector that captures only the complementary physical information. To achieve this compact joint encoding, we draw inspiration from video compression, where 3D variational autoencoders (VAEs)~\cite{wan2025wan,kong2024hunyuanvideo,peng2025open,zhao2024cv} model dependencies across temporally coherent frames. We note a natural analogy: just as consecutive video frames share temporal coherence, RGB appearance and PBR maps are physically coupled through shared material properties. 

We formulate the RGB and four PBR maps as a unified 5-frame input, enabling us to adapt video diffusion architectures to learn their joint distribution via a 3D VAE.

Building on this joint representation, MatPedia employs a video Diffusion 
Transformer (DiT) operating on the joint latent space to perform diverse 
material generation tasks—including text-to-material, image-to-material, and 
intrinsic decomposition—by conditioning on different input modalities. We 
initialize the DiT from pre-trained video generation models~\cite{wan2025wan,kong2024hunyuanvideo,peng2025open,zhao2024cv} 
and adapt it via LoRA~\cite{hu2022lora}, transferring visual priors from 
large-scale RGB data.
To train MatPedia, we construct \textbf{MatHybrid-410K}, a mixed corpus 
combining open-source PBR datasets~\cite{vecchio2024matsynth,ma2023opensvbrdf,deschaintre2018single} 
with a large-scale RGB image collection. Beyond standard training on RGB-PBR paired 
data, we incorporate abundant RGB-only images to enhance the model's 
generalization capability and generation quality, enabling it to leverage 
diverse visual knowledge from the broader image domain. 
By jointly modeling multiple material representations and unifying diverse 
material-related tasks within a single architecture, MatPedia—named for its 
encyclopedic coverage of material understanding and generation—achieves 
native 1024×1024 generation across a wide range of material types.

In summary, our main contributions are:

\begin{itemize}

\item We propose a \textbf{joint RGB-PBR representation} that enables unified 
material modeling while leveraging large-scale RGB image data.

\item We present \textbf{MatPedia}, a unified framework handling multiple 
material tasks—text-to-material, image-to-material, and intrinsic decomposition—
achieving native 1024×1024 generation.

\item We construct and will release \textbf{MatHybrid-410K}, a large-scale mixed dataset designed to facilitate a hybrid training strategy, enhancing generation quality and diversity.

\end{itemize}

%% file: sec/2_related_work.tex
\section{Related Work}
\label{sec: related work}

\begin{figure*}[htbp]
    \centering
    \includegraphics[width=1\linewidth]
    {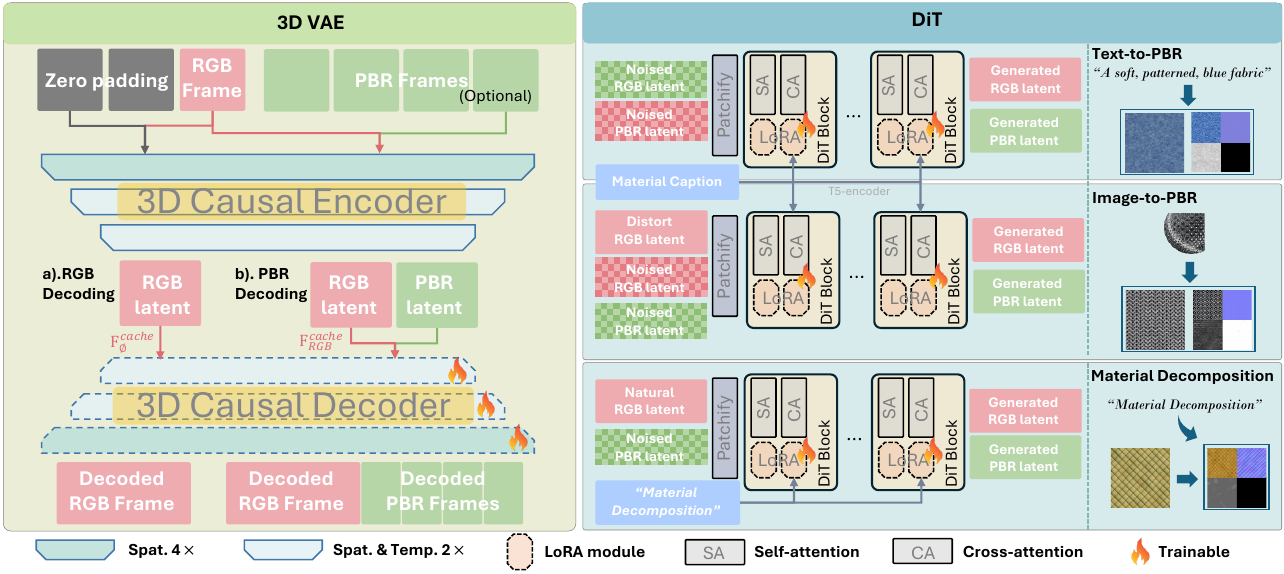}
    \vspace{-20pt}
    \caption{Pipeline of the proposed MatPedia framework. \textbf{Left:} The \textbf{3D VAE} encodes a shaded RGB frame together with optional PBR maps into a \textit{joint RGB-PBR latent representation}, where PBR maps are conditioned on the RGB appearance. This compact representation supports both (a) shaded RGB decoding and (b) PBR decoding at native $1024\times1024$ resolution. \textbf{Right:} The \textbf{DiT}, initialized from large-scale video generation models and adapted via LoRA, operates on the joint latents to perform three tasks: \textit{Text-to-PBR} (generate RGB/PBR from material captions), \textit{Image-to-PBR} (generate planar RGB/PBR from distorted input images), and \textit{Material Decomposition} (recover PBR maps from natural images). DiT blocks integrate self-attention (SA), cross-attention (CA), and LoRA modules to enable flexible conditioning across modalities.}  
    \vspace{-14pt}
    \label{fig:pipeline}  
\end{figure*}

\subsection{Material Estimation and Generation}
Recovering physically-based material representations from images is a long-standing challenge in computer graphics~\cite{guarnera2016brdf}. Traditional methods often rely on strong priors or controlled capture setups, such as known illumination~\cite{chandraker2014shape,hui2015dictionary,riviere2014mobile}, self-similarity~\cite{aittala2015two}, or specific spatial relations between views~\cite{xu2016minimal}, which limit applicability in uncontrolled environments. With the introduction of large synthetic datasets~\cite{deschaintre2018single}, deep learning methods are applied to material recovery from single or multiple images~\cite{guo2021highlight,shi2020match,zhou2022look,vecchio2021surfacenet,luo2024correlation,xing2025diffusion}. While these approaches relax some capture constraints, many still require near fronto-parallel views or specific lighting, making in-the-wild material extraction challenging.

Generative models offer an alternative via learned priors for material synthesis. GAN-based methods~\cite{guo2020materialgan,zhou2022tilegen,zhou2023photomat} enable unconditional generation and latent optimization, while diffusion models~\cite{rombach2022high, sartor2023matfusion} improve stability and controllability. Text2Mat~\cite{he2023text2mat} generates PBR materials from text using a Stable Diffusion-based model~\cite{rombach2022high}, and ReflectanceFusion~\cite{xue2024reflectancefusion} produces editable PBR maps via tandem diffusion, but both face "baked lighting" issues. ControlMat~\cite{vecchio2024controlmat} synthesizes tileable high-res PBR materials from single photos via diffusion, but may have slight appearance deviations. MatFuse~\cite{vecchio2024matfuse} unifies multi-modal material creation and editing but lacks diversity due to small training datasets.
Material Palette~\cite{lopes2024material} extracts tileable PBR materials from real images but struggles with uniform materials, illumination ambiguities, and strong geometric distortions including wrinkled or distorted surfaces. RGB$\leftrightarrow$X~\cite{zeng2024rgb} synthesize RGB from specified channels and estimate intrinsic channels from images and via modality switching, but this incurs longer inference time during material decomposition. 
MaterialPicker~\cite{ma2024materialpicker} handles such challenging images via a video backbone but compresses frames independently (limiting resolution to $256\times256$). Regarding IntrinsicX~\cite{kocsis2025intrinsix}: they use per-channel LoRAs (one per PBR map) with cross-attention for consistency, while ours uses per-task LoRAs on a shared joint latent space. We adopt a joint RGB-PBR representation that encodes PBR maps conditioned on the RGB appearance, achieving a high compression ratio while preserving material detail, and supporting native $1024\times1024$ resolution generation.

\subsection{Video Diffusion Models}

Video diffusion models extend denoising frameworks to spatio-temporal data, enabling coherent motion modeling and high-quality video synthesis. Early works such as Imagen Video~\cite{ho2022imagen} and Phenaki~\cite{villegas2022phenaki} introduced latent-space spatio-temporal attention and causal modeling for text-conditioned and long-form generation. Subsequent systems, including Video LDM~\cite{blattmann2023align} and Sora~\cite{brooks2024video}, improved multi-modal alignment, scalability, and fidelity, achieving high-resolution, long-duration videos with complex dynamics. Specialized designs like CogVideo~\cite{hong2022cogvideo} and Pyramid Video Diffusion~\cite{jin2024pyramidal} further enhanced efficiency and detail through hybrid attention and hierarchical latent spaces. Spatio-temporal VAEs are central to these pipelines, providing compact video representations while preserving temporal coherence, as demonstrated by CogVideoX~\cite{yang2024cogvideox}, HunyuanVideo~\cite{kong2024hunyuanvideo}, VideoGen~\cite{li2025wf}, and Improved ST-VAE~\cite{wu2025improved}. Wan~\cite{wan2025wan} introduces a 3D VAE with causal convolutions and high compression ratios, delivering efficient text-and-image-to-video synthesis with strong temporal consistency, which we adopt for our joint RGB-PBR representation.

%% file: sec/3_method.tex
\section{Preliminaries}
\label{sec:preliminary}

\noindent\textbf{Spatio-Temporal Variational Autoencoder.}  
Variational Autoencoders (VAEs)~\cite{kingma2013auto,blattmann2023align,wan2025wan} map high-dimensional video data to a compact latent space for efficient generative modeling. The Wan2.2-VAE~\cite{wan2025wan} adopts a 3D causal convolution design with high spatio-temporal compression: the first frame is only downsampled by $16\times$ spatially, while the remaining $T$ frames are downsampled by $16\times$ spatially and $4\times$ temporally, producing a latent tensor of shape $[\,1+T/4,\,H/16,\,W/16\,]$.

\noindent\textbf{Material Representation.}  
We represent each material as a Spatially Varying Bidirectional Reflectance
Distribution Function parameterized 
by four maps: basecolor $\mathbf{a}$ (diffuse albedo), normal $\mathbf{n}$ 
(surface orientation), roughness $\mathbf{r}$ (microfacet distribution width), 
and metallic $\mathbf{m}$ (metallic factor). Together, these maps encode the 
material's appearance under varying lighting and viewing conditions following 
the Cook-Torrance microfacet model~\cite{cook1982reflectance,walter2007microfacet}.

\section{Method}
\label{sec:method}

\subsection{Overview}

Our goal is to develop a unified high-resolution generative model (Fig.~\ref{fig:pipeline}) that integrates three PBR material tasks—text-to-material generation, image-to-material generation, and intrinsic decomposition—within a single architecture.
At its core is a joint RGB-PBR representation (Sec.~\ref{sec:3d vae}): a fine-tuned 3D VAE encodes materials into two interdependent latents, one for shaded RGB appearance and one jointly encoding four PBR maps. Building on this, we adopt a video DiT backbone with LoRA fine-tuning (Sec.~\ref{sec:GMM}) that handles all three tasks through flexible conditioning. We curate a large-scale hybrid dataset (Sec.~\ref{sec:dataset_text}) combining RGB-PBR pairs and RGB-only samples for training.

\subsection{Joint RGB-PBR Representation}
\label{sec:3d vae}

Our method builds on a \textit{joint RGB-PBR representation}: one latent 
encodes the shaded RGB appearance, and another jointly encodes all four PBR 
maps. Crucially, RGB appearance already captures substantial visual information 
(texture, color, structure), while PBR maps primarily encode complementary 
physical properties (surface geometry, material type, reflectance). This 
complementary structure enables highly compact encoding—the PBR latent need 
only represent incremental physical attributes rather than redundantly encoding 
visual structure already present in RGB.

Formally, given an RGB image $\mathbf{I}_{\text{rgb}} \in \mathbb{R}^{H \times W \times 3}$ 
and four PBR maps (basecolor $\mathbf{a}$, normal $\mathbf{n}$, roughness 
$\mathbf{r}$, metallic $\mathbf{m}$), we employ a pretrained video 
VAE~\cite{wan2025wan} that treats them as a 5-frame sequence along the 
temporal dimension. Video VAEs exploit redundancies across frames through 
3D convolutions, naturally capturing the structural dependencies between 
RGB and PBR. The encoder produces two interdependent latents:
\begin{equation}
\mathbf{z}_{\text{rgb}} = \mathcal{E}_{\text{rgb}}(\mathbf{I}_{\text{rgb}}), \,\,\, \mathbf{z}_{\text{pbr}} = \mathcal{E}_{\text{pbr}}([\mathcal{F}_{\text{enc}}(\mathbf{z}_{\text{rgb}}), \mathbf{a}, \mathbf{n}, \mathbf{r}, \mathbf{m}]),
\end{equation}
where RGB is encoded independently, while PBR maps are encoded using cached 
features $\mathcal{F}_{\text{enc}}$ from the RGB encoding branch, exploiting 
their inherent correlation for compact representation.

The decoder mirrors this asymmetric structure:
\begin{equation}
\mathbf{I}'_{\text{rgb}} = \mathcal{D}_{\text{rgb}}(\mathbf{z}_{\text{rgb}}), \,\,\, \{\mathbf{a}', \mathbf{n}', \mathbf{r}', \mathbf{m}'\} = \mathcal{D}_{\text{pbr}}([\mathcal{F}_{\text{dec}}(\mathbf{z}_{\text{rgb}}), \mathbf{z}_{\text{pbr}}]),
\end{equation}
where RGB latents are decoded independently, while PBR latents are decoded 
using cached features $\mathcal{F}_{\text{dec}}$ from the RGB decoder, 
producing high-fidelity PBR maps as incremental refinements. This symmetric 
encoder-decoder design, combined with compact dual-latent encoding, enables 
scalable high-resolution synthesis (Fig.~\ref{fig:pipeline}).

While the pretrained video VAE provides strong visual priors from large-scale 
data, we fine-tune the decoder on PBR material data to achieve high-fidelity 
material reconstruction, keeping the encoder fixed to preserve the pretrained 
latent distribution. The decoder is optimized with pixel-wise and perceptual 
losses:
\begin{equation}
    \mathcal{L}_{\mathrm{VAE}} = \lambda_1 \left\| \hat{\mathbf{x}} - \mathbf{x} \right\|_1 
    + \lambda_2 \left\| \phi(\hat{\mathbf{x}}) - \phi(\mathbf{x}) \right\|_2^2,
\end{equation}
where $\mathbf{x}$ denotes the ground-truth 5-frame sequence, $\hat{\mathbf{x}}$ 
is the reconstruction, and $\phi(\cdot)$ extracts features from a pretrained 
VGG network~\cite{simonyan2014very}.

\subsection{Universal Generative Material Model}
\label{sec:GMM}

Building on our joint RGB-PBR representation, we adopt a video DiT~\cite{wan2025wan} as the generative backbone. The model treats 
materials as two interdependent latents: one for RGB appearance and one jointly 
encoding the four PBR maps. By leveraging video priors, the DiT naturally 
captures cross-map correlations while maintaining spatial alignment. We unify 
three PBR-related tasks within this single architecture through flexible 
conditioning and LoRA fine-tuning~\cite{hu2022lora} (as shown in 
Fig.~\ref{fig:pipeline}), enabling diverse material generation at 1024×1024 
resolution.

\noindent\textbf{Text-to-Material Generation.}  
The DiT generates both RGB and PBR latents from noise conditioned on text 
prompts. The decoder then reconstructs the complete material: the RGB latent 
is decoded independently, and the PBR latent is decoded with feature caching 
from the RGB decoder, producing spatially aligned RGB appearance and PBR maps 
at 1024×1024 resolution. Crucially, we train this task using MatHybrid-410K, 
which includes both RGB-PBR paired sequences and RGB-only images. For RGB-only 
samples, we supervise only the RGB latent generation, allowing the model to 
learn rich visual priors from large-scale RGB data without requiring 
corresponding PBR annotations. Importantly, this strategy keeps the PBR latent 
distribution intact, as it continues to be learned from RGB-PBR paired data, 
while the RGB branch benefits from the additional diverse visual knowledge.

\noindent\textbf{Image-to-Material Generation.}  
Given a potentially distorted RGB photograph, we encode it through the VAE's 
encoder as a conditioning latent. The DiT then generates two new latents—
rectified planar RGB and PBR—conditioned on this encoding. During decoding, 
the generated RGB latent is reconstructed independently (rectifying geometric 
distortions), while the PBR latent is decoded with cached RGB features, 
producing flat material representations with accurate physical properties. 
This task is fine-tuned from the text-to-material checkpoint using LoRA, 
transferring the learned RGB priors to image-conditioned generation.

\begin{figure}[t]
    \centering
    \includegraphics[width=1\linewidth]{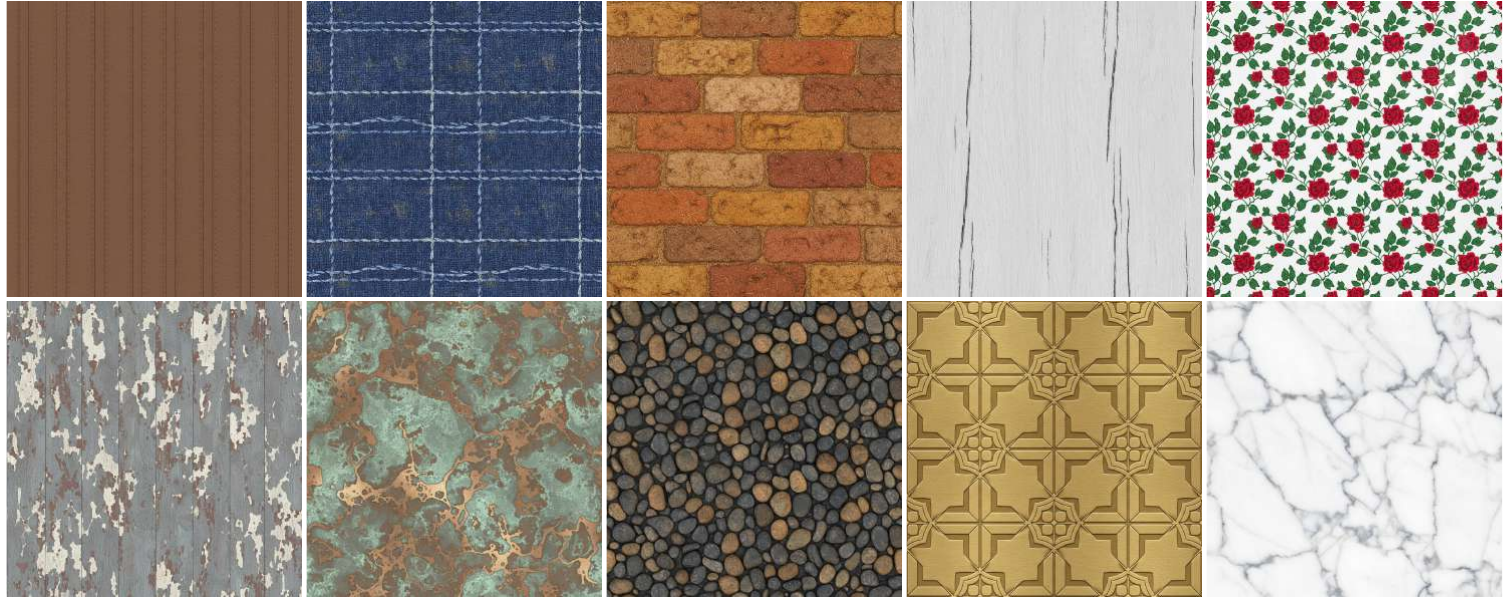}  
    \vspace{-15pt}
    \caption{Examples of planar material images from the RGB Appearance Dataset, generated by the Gemini~2.5~Flash~Image model~\cite{gemini2025flash}.}
    \label{fig:dataset}  
    \vspace{-15pt}
\end{figure}

\noindent\textbf{Intrinsic Decomposition.}  
Given a planar RGB image, we encode it into an RGB latent and condition the 
DiT to generate only the corresponding PBR latent. The decoder then reconstructs 
the PBR maps using feature caching from the encoded RGB latent, decomposing 
the rendered appearance into underlying physical properties while ensuring 
spatial alignment. Similar to image-to-material generation, this task is 
fine-tuned from the text-to-material weights via LoRA, enabling the model to 
leverage RGB visual knowledge for accurate material decomposition.

\noindent\textbf{Training Strategy.}  
All tasks are optimized using the rectified flow objective~\cite{liu2022flow}:

\begin{equation}
    \mathcal{L}_{\mathrm{RF}} = \mathbb{E}_{\mathbf{x}_0, \mathbf{x}_1, t} 
    \left[ \left\| v_\theta\left( \mathbf{x}_t, t, \mathbf{c} \right) - 
    \left( \mathbf{x}_0 - \mathbf{x}_1 \right) \right\|_2^2 \right],
\end{equation}
where $\mathbf{x}_0$ denotes the ground-truth latent, $\mathbf{x}_1 \sim 
\mathcal{N}(\mathbf{0}, \mathbf{I})$ is sampled noise, $\mathbf{x}_t = 
(1-t)\,\mathbf{x}_0 + t\,\mathbf{x}_1$ is the interpolated latent at timestep 
$t$, $\mathbf{c}$ is the conditioning signal (text or image features), and 
$v_\theta$ is the velocity prediction network. For text-to-material generation, 
the loss is applied to both RGB and PBR latents for paired data, or solely to 
the RGB latent for RGB-only samples. Image-to-material and intrinsic 
decomposition are fine-tuned from this text-to-material checkpoint via LoRA.

\subsection{Dataset}
\label{sec:dataset_text}

\label{sec:dataset}

\begin{table}[t]
  \centering
  \caption{Quantitative comparison of text-to-material generation using CLIP and DINO-FID metrics. Best results are in \textbf{bold}.}
  \vspace{-8pt}
    \resizebox{\linewidth}{!}{
    \begin{tabular}{c|cc|c}
    \toprule
          & MatFuse & Ours  & Ours(w/o mixed training) \\
    \midrule
    CLIP$\uparrow$  & 0.261 & \textbf{0.283} & 0.275 \\
    DINO-FID$\downarrow$ & 1.90   & \textbf{1.31} & 1.62 \\
    \bottomrule
    \end{tabular}%
    }
    \vspace{-15pt}
  \label{tab:text_pbr_metrics}%
\end{table}

To support diverse material generation and robust material recovery, we construct MatHybrid-410K, a large-scale hybrid dataset with two complementary subsets.

\noindent\textbf{RGB Appearance Dataset.}  
We collect approximately 50,000 planar material images from two sources: (1) procedurally generated flat surfaces (as shown in Fig.~\ref{fig:dataset}) using Gemini~2.5~Flash~Image~\cite{gemini2025flash}, and (2) real-world planar material photographs from public repositories. Each image is paired with a text description generated by Qwen2.5-VL-72B-Instruct~\cite{bai2025qwen2}, enabling text-to-material generation. This RGB-only subset provides diverse appearance patterns for training without requiring paired PBR maps.

\noindent\textbf{Complete PBR Material Dataset.}  
We source approximately 6,000 complete PBR material sets from Matsynth~\cite{vecchio2024matsynth} and additional collections. For each material, we render two types of training data using Blender's Disney Principled BSDF~\cite{mcauley2012practical}: (1) planar views under 32 HDR environment maps, yielding 192,000 pairs for intrinsic decomposition training; (2) distorted views rendered on geometric primitives (cubes, spheres, cylinders, cones, toruses) with varied lighting and camera angles, yielding approximately 168,000 pairs for image-to-material generation. Distorted images are cropped to ensure the material occupies at least 70\% of the area.

\subsection{Training and Inference}
\label{sec:Train}

\noindent\textbf{3D VAE Training.}  
We fine-tune the decoder on RGB-PBR paired data (Sec.~\ref{sec:dataset_text}) at 1024×1024 for 10K steps using AdamW (lr=$5\times10^{-5}$, $\lambda_1=10$, $\lambda_2=1$).

\noindent\textbf{DiT Training.}  
We train the video DiT~\cite{wan2025wan} with LoRA~\cite{hu2022lora} (rank 128) on attention projections and FFN linear layers. Each task uses the hybrid dataset at 1024×1024 resolution, trained for 200K steps with batch size 16 and learning rate $1\times10^{-4}$.

\noindent\textbf{Inference.}  
Generation at 1024×1024 is upsampled to 4K using RealESRGAN~\cite{wang2021realesrgan}. Complete PBR map generation takes 20s with 50 sampling steps.

%% file: sec/4_experiment.tex
\section{Experiments}

We evaluate our method on the same test set as MaterialPicker~\cite{ma2024materialpicker}.  
Our experiments are organized into four parts: text-to-material generation, image-to-material generation, material decomposition, and ablation studies.  
We adopt the following evaluation metrics:  
\textbf{CLIP score}~\cite{radford2021learning} measures semantic alignment, evaluating \emph{text–image} similarity for text-conditioned generation and \emph{image–image} similarity for image-conditioned generation.  
\textbf{DINO score}~\cite{stein2023exposing} measures perceptual similarity between generated and reference images using DINOv2~\cite{oquab2023dinov2} embeddings.  
\textbf{DINO-FID} extends the Fréchet Inception Distance~\cite{heusel2017gans} by replacing Inception features with DINOv2 embeddings, providing perceptually relevant distribution similarity.  
\textbf{MSE} measures pixel-wise reconstruction error.  
\textbf{LPIPS}~\cite{zhang2018unreasonable} quantifies perceptual similarity using deep feature distances, correlating well with human judgments.  
For all metrics, higher CLIP and DINO scores indicate better alignment, while lower DINO-FID, MSE, and LPIPS indicate better fidelity. More results are provided in the supplementary.

\subsection{Text-to-Material Generation}

\begin{table}[t]
  \centering
  \caption{Quantitative comparison of image-to-material generation using CLIP score and DINO score metrics. Best results are in \textbf{bold}.}
  \vspace{-8pt}
    \resizebox{\linewidth}{!}{
    \begin{tabular}{c|cccc}
    \toprule
    \textbf{CLIP score$\uparrow$} & basecolor & Normal & Roughness & Render \\
    \midrule
    MatFuse & 0.833 & 0.906 & 0.873 & 0.859 \\
    Material Palette & 0.813 & 0.875 & 0.780 & 0.824 \\
    Ours  & \textbf{0.943} & \textbf{0.927} & \textbf{0.903} & \textbf{0.923} \\
    \midrule
    \textbf{DINO score$\uparrow$} & basecolor & Normal & Roughness & Render \\
    \midrule
    MatFuse & 0.649 & 0.755 & 0.717 & 0.677 \\
    Material Palette & 0.543 & 0.579 & 0.448 & 0.541 \\
    Ours  & \textbf{0.907} & \textbf{0.762} & \textbf{0.752} & \textbf{0.843} \\
    \bottomrule
    \end{tabular}%
    }
    \vspace{-14pt}
  \label{tab:distored-image}%
\end{table}

Since MaterialPicker~\cite{ma2024materialpicker} and ControlMat~\cite{vecchio2024controlmat} are not publicly available, we perform qualitative comparisons with these methods using the cases presented in the MaterialPicker paper. Quantitative evaluation is conducted only against MatFuse~\cite{vecchio2024matfuse}, a unified diffusion-based framework supporting multimodal material creation and editing, using CLIP score (text–image similarity) and DINO-FID (distribution similarity).

\begin{figure}[t]
    \centering
    \includegraphics[width=1\linewidth]{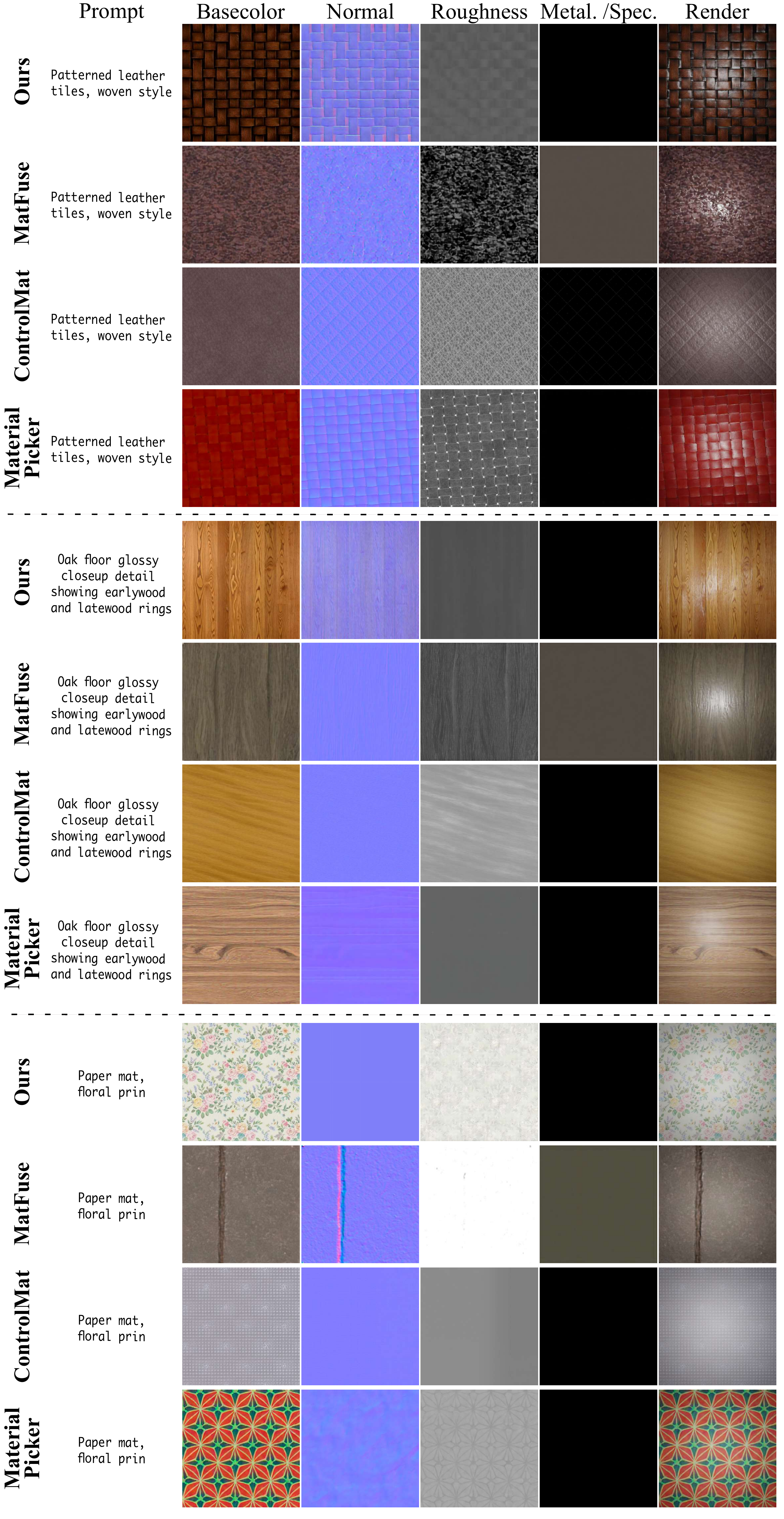}
    \vspace{-10pt}
    \caption{Qualitative comparison of text-conditioned PBR material generation among our method, MatFuse~\cite{vecchio2024matfuse}, ControlMat~\cite{vecchio2024controlmat}, and MaterialPicker~\cite{ma2024materialpicker}. For each prompt, we show the generated PBR maps (Basecolor, Normal, Roughness, Metallic) followed by a render view under point-light illumination. We note that MatFuse generates a specular map rather than a metallic map.}
    \label{fig:text_pbr_com}
    \vspace{-20pt}
\end{figure}

As shown in Table~\ref{tab:text_pbr_metrics}, our method achieves higher 
CLIP score and lower DINO-FID than MatFuse, indicating better semantic 
consistency and visual fidelity. The qualitative comparison in 
Fig.~\ref{fig:text_pbr_com} further demonstrates that our method produces 
materials with more accurate patterns, material attributes, and surface 
details. For instance, in the "Oak floor glossy" case, our method correctly 
captures both the wood grain structure and realistic gloss distribution, 
while baseline methods struggle with either texture clarity or material properties.

\subsection{Image-to-Material Generation}

We compare our method against MatFuse and Material Palette~\cite{lopes2024material}, which extract PBR materials from single real-world images via diffusion-based texture synthesis and PBR decomposition.  
Evaluation uses CLIP score (image–image similarity) and DINO score (perceptual similarity).

\begin{figure}[t]
    \centering
    \includegraphics[width=1\linewidth]{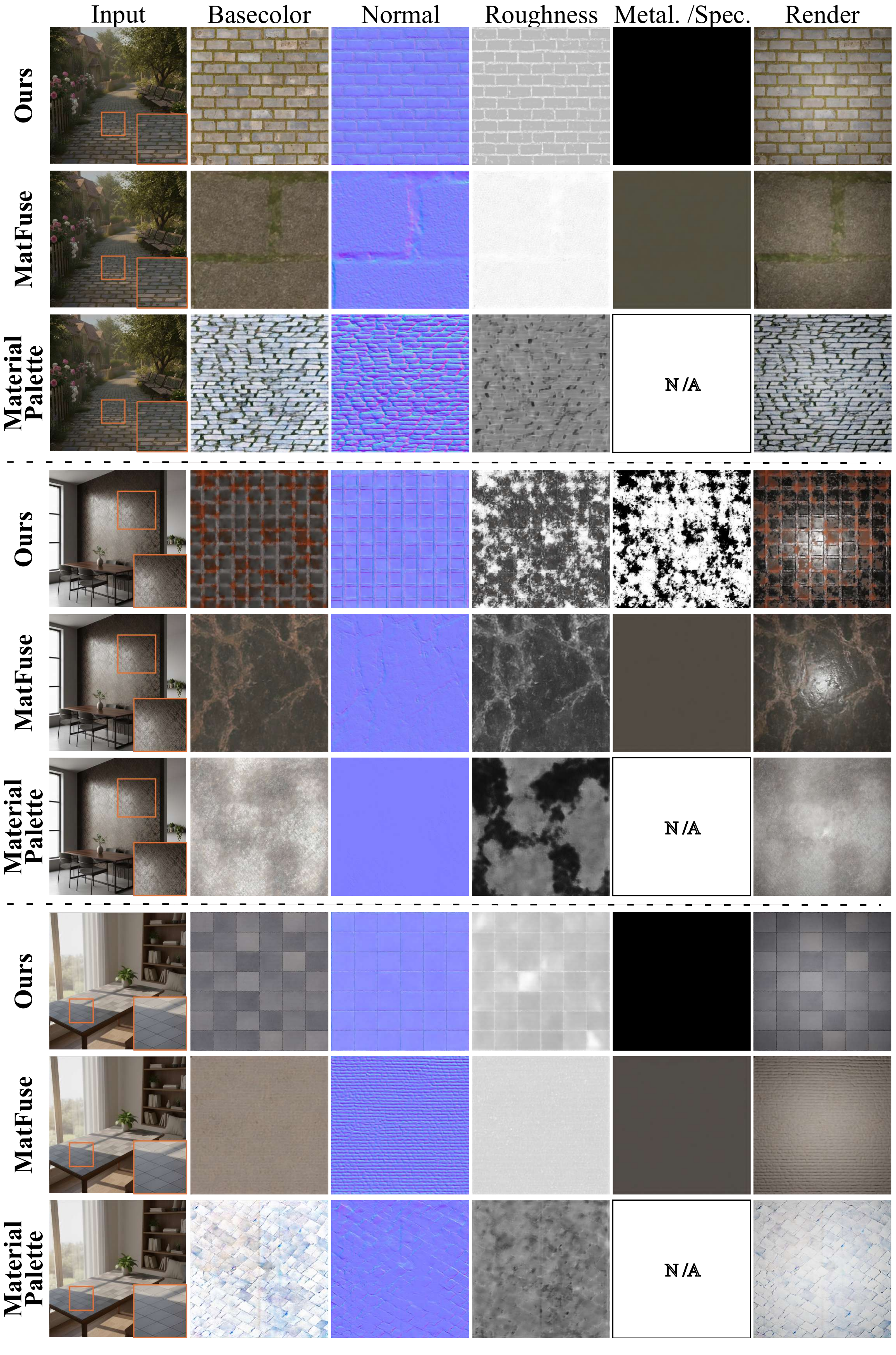}
    \vspace{-15pt}
    \caption{Qualitative comparison of image-conditioned PBR generation. For each sample, the first column shows the distorted input image (cropped from the scene), and the second to last columns present the generated material maps together with a rendering under point-light illumination. Our method produces geometrically flattened and artifact-free maps, while MatFuse shows reduced roughness fidelity and Material Palette retains geometric distortions from the input. }
    \vspace{-15pt}
    \label{fig:distord_com}
\end{figure}

Our method achieves the highest CLIP and DINO scores across all channels (Table~\ref{tab:distored-image}), with notable gains in \emph{basecolor} (+0.11 CLIP, +0.26 DINO over MatFuse), indicating improved recovery of intrinsic colors under distortion.  
Qualitatively (Fig.~\ref{fig:distord_com}), our framework generates flattened, artifact-free basecolor, normal, and roughness maps with fine detail and spatial consistency.  
In contrast, MatFuse exhibits muted texture detail and reduced roughness fidelity, while Material Palette fails to remove geometric distortions, leaving shading and perspective artifacts in the generated maps.

\subsection{Material Decomposition}

\begin{figure}[t]
    \centering
    \includegraphics[width=1\linewidth]{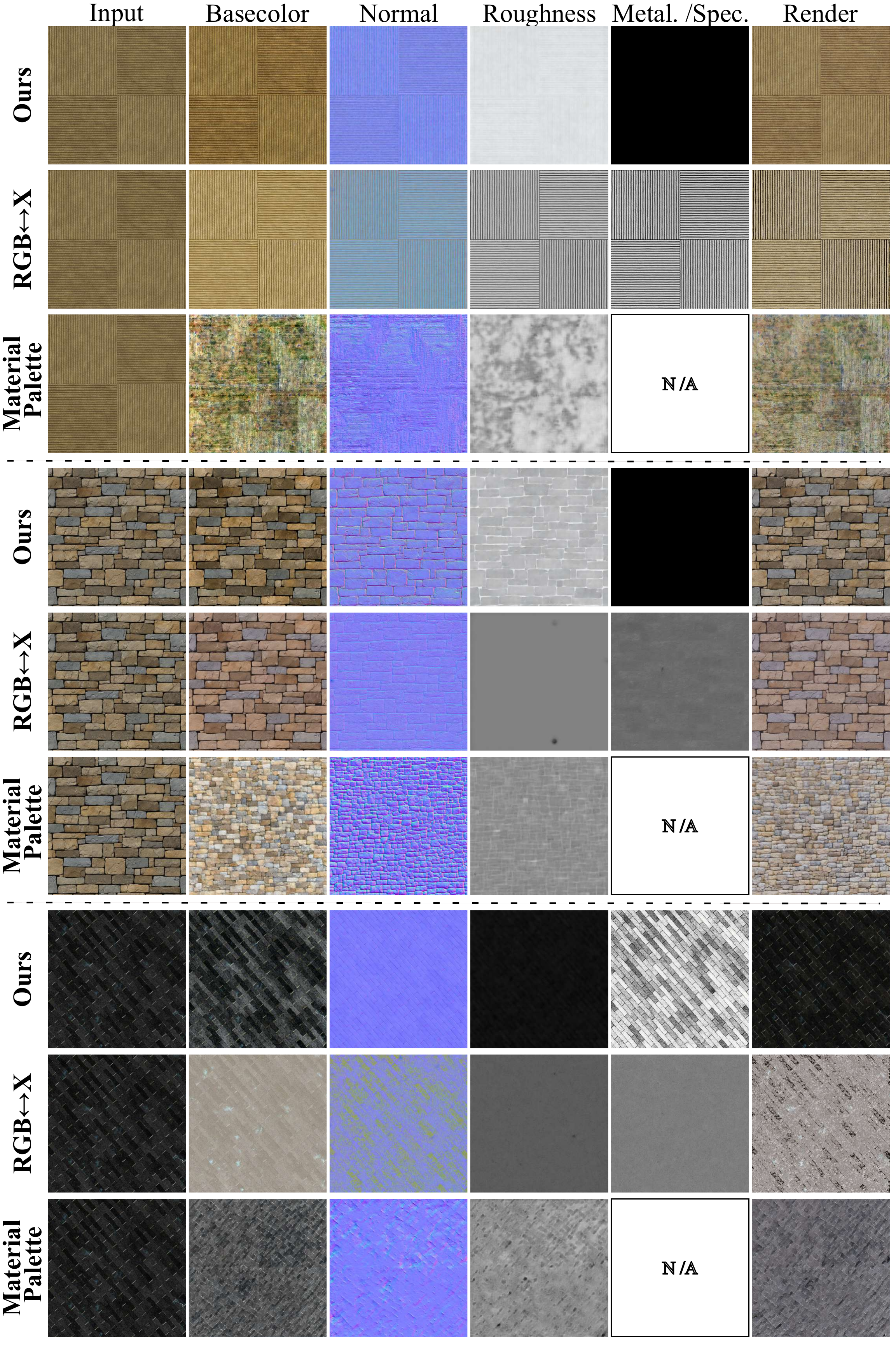}
    \vspace{-15pt}
    \caption{Qualitative comparison of material decomposition. For each sample, the first column shows the planar input image, and the second to last columns present the generated material maps together with a rendering under environment lighting. Our method produces consistent structural patterns, yielding rendered views that closely match the input appearance.}
    \label{fig:image_pbr_com}
    \vspace{-5pt}
\end{figure}

\begin{table}[htbp]
  \centering
  \vspace{-5pt}
  \caption{Quantitative comparison of material decomposition using MSE and LPIPS metrics. Best results are in \textbf{bold}.}
  \vspace{-5pt}
    \resizebox{\linewidth}{!}{
    \begin{tabular}{c|cccc}
    \toprule
    \textbf{MSE$\downarrow$} & \textbf{basecolor} & \textbf{Normal} & \textbf{Roughness} & \textbf{Render} \\
    \midrule
    Material Palette & 0.058 & 0.014 & 0.103 & 0.034 \\
    RGB$\leftrightarrow$X & 0.122 & 0.034 & 0.083 & 0.079 \\
    Ours  & \textbf{0.009} & \textbf{0.007} & \textbf{0.041} & \textbf{0.008} \\
    \midrule
    \textbf{LPIPS$\downarrow$} & \textbf{basecolor} & \textbf{Normal} & \textbf{Roughness} & \textbf{Render} \\
    \midrule
    Material Palette & 0.716 & 0.513 & 0.761 & 0.644 \\
    RGB$\leftrightarrow$X & 0.770 & 0.643 & 0.622 & 0.706 \\
    Ours  & \textbf{0.664} & \textbf{0.419} & \textbf{0.423} & \textbf{0.627} \\
    \bottomrule
    \end{tabular}%
    }
    \vspace{-15pt}
  \label{tab:image_pbr_metrics}%
\end{table}%

We compare our method against Material Palette~\cite{lopes2024material} and RGB$\leftrightarrow$X~\cite{zeng2024rgb}. Evaluation uses MSE (pixel-wise error) and LPIPS (deep feature perceptual distance) for reconstruction quality assessment.

Table~\ref{tab:image_pbr_metrics} and Fig.~\ref{fig:image_pbr_com} compare our method with Material Palette and RGB$\leftrightarrow$X for PBR material decomposition from single RGB inputs. Across diverse material categories, our approach consistently achieves the lowest MSE and LPIPS, indicating more accurate pixel-level reconstruction of PBR maps. Qualitatively, it produces basecolor maps with accurate colors and minimal noise, normal maps with sharp geometric details, and roughness maps with coherent structural patterns, yielding rendered views that closely match the inputs. In contrast, RGB$\leftrightarrow$X and Material Palette exhibit artifacts, loss of fine details, and tonal inconsistencies, especially in challenging textures such as woven surfaces and irregular stone.

\subsection{Ablation Study}

\paragraph{3D VAE Decoder Fine-tuning.}
\begin{figure}[t]
    \centering
    \includegraphics[width=1\linewidth]{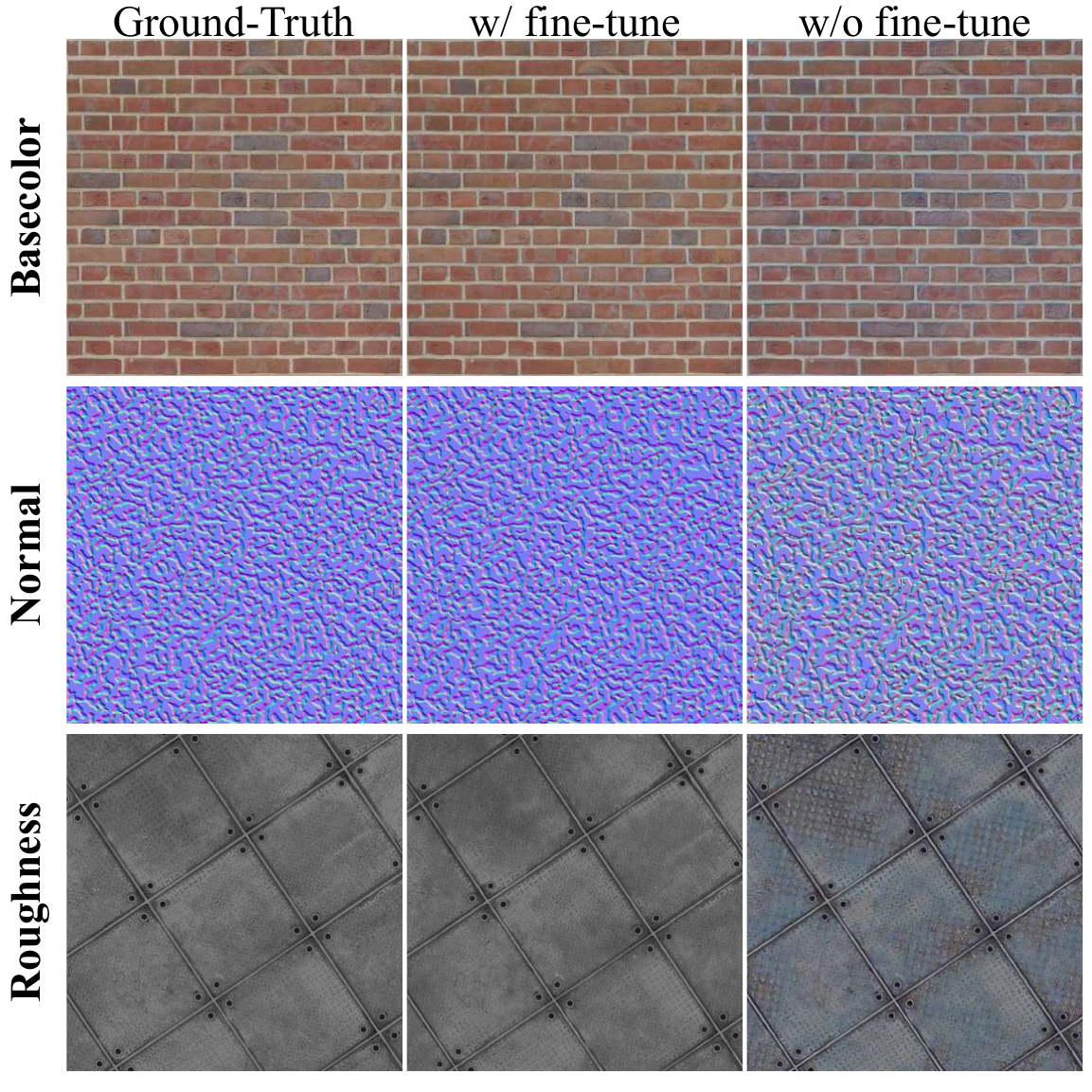}
    \vspace{-18pt}
    \caption{Visual comparison of VAE reconstruction quality before and after decoder fine-tuning. Fine-tuning significantly improves fidelity for basecolor, normal, and roughness maps, with sharper details and reduced artifacts.}
    \label{fig:vae}
    \vspace{-15pt}
\end{figure}

Table~\ref{tab:vae_psnr} shows that decoder fine-tuning substantially improves 
reconstruction quality, with the largest gains in \emph{Normal} (+3.55~dB) and 
\emph{Roughness} (+5.20~dB), both critical for material appearance. 
Fig.~\ref{fig:vae} confirms sharper reconstructions with fewer artifacts.

\begin{table}[htbp]
  \centering
  \caption{Reconstruction quality (PSNR) before and after decoder fine-tuning on our PBR dataset. Fine-tuning improves fidelity across all channels.}
  \vspace{-5pt}
  \resizebox{\linewidth}{!}{
    \begin{tabular}{l|ccccc}
    \toprule
    Method & Render & Basecolor & Normal & Roughness & Metallic \\
    \midrule
    Before fine-tuning & 33.12 & 29.27 & 27.29 & 31.36 & 42.45 \\
    After fine-tuning & \textbf{34.25} & \textbf{31.30} & \textbf{30.84} & \textbf{36.56} & \textbf{45.59} \\
    \bottomrule
    \end{tabular}
  }
  \label{tab:vae_psnr}
  \vspace{-22pt}
\end{table}

\paragraph{Effect of Hybrid Training.}

We evaluate the impact of hybrid training by comparing our full model with a variant trained solely on the Complete PBR Material Dataset. Table~\ref{tab:text_pbr_metrics} shows that excluding the RGB Appearance Dataset degrades performance: CLIP score drops from $0.283$ to $0.275$, and DINO-FID increases from $1.31$ to $1.62$. This confirms that incorporating the RGB Appearance Dataset enhances both semantic alignment and perceptual realism in text-conditioned generation.

\paragraph{Generalization to Point-Light Illumination.}
\begin{figure}[t]
    \centering
    \includegraphics[width=1\linewidth]{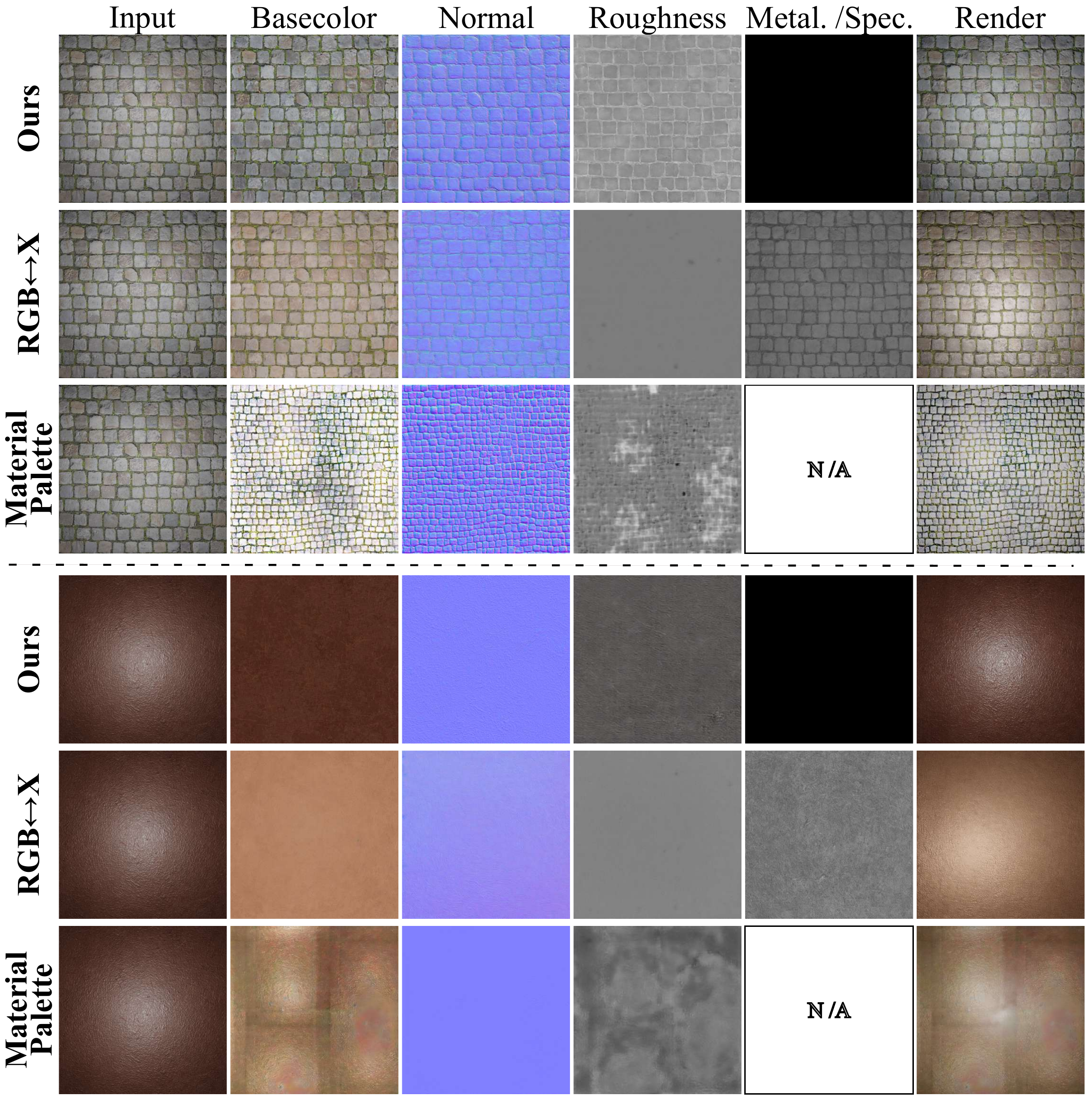}
    \vspace{-15pt}
    \caption{Qualitative comparison under point-light illumination, which introduces strong shading and non-uniform lighting absent from training data.}
    \label{fig:point_image_pbr_com}
    \vspace{-15pt}
\end{figure}

We evaluate material extraction from inputs captured under point-light 
illumination, which introduces strong shading and non-uniform lighting 
patterns. As shown in Fig.~\ref{fig:point_image_pbr_com}, our method produces basecolor, normal, and roughness maps whose rendered appearance closely matches the input image, indicating effective disentanglement even under challenging illumination.
In contrast, RGB$\leftrightarrow$X shows color shifts and reduced detail 
fidelity, while Material Palette exhibits severe artifacts in the extracted 
maps.

\section{Conclusion and Discussion}
We introduce MatPedia, a unified PBR material generation framework built on a novel joint RGB-PBR representation. By treating materials as a 5-frame sequence and leveraging a video VAE, our approach captures cross-map correlations while enabling the use of abundant RGB data to overcome PBR data scarcity. MatPedia unifies three material-related tasks—text-to-material generation, image-to-material generation, and intrinsic decomposition—within a single architecture, producing high-fidelity results at $1024\times1024$ resolution that surpass existing methods in quality, diversity, and scalability. While the joint compression approach couples spatial features, limiting direct support for tileable generation via noise rolling, the native $1024\times1024$ resolution (upsampled to $4096\times4096$) provides sufficiently large textures for most production scenarios. Future work will focus on expanding to additional material channels (e.g., height, subsurface scattering) to enable more comprehensive physical modeling for advanced rendering pipelines.

\section*{Acknowledgment}
We thank the reviewers for the valuable comments. This work has been partially supported by the National Natural Science Foundation of China under grant No. 62572230.